\pdfoutput=1
\documentclass[10pt,twocolumn,letterpaper]{article}

\usepackage{3dv}
\usepackage{authblk}
\usepackage{times}
\usepackage{epsfig}
\usepackage{graphicx}
\usepackage{amsmath}
\usepackage{amssymb}
\usepackage[dvipsnames]{xcolor}
\usepackage{dsfont}
\usepackage{booktabs}
\usepackage{stfloats}

\newcommand{\str}[1]{\ \textcolor{ForestGreen}{\textbf{STR:} #1}}

\usepackage[pagebackref=true,breaklinks=true,colorlinks,bookmarks=false]{hyperref}

\threedvfinalcopy 

\def\threedvPaperID{61} 

\usepackage[normalem]{ulem}

\newcommand{\newcolor}{\color{BrickRed}}
\newcommand{\oldcolor}{\color{gray}}
\newcommand{\new}[1]{{\newcolor#1}}
\newcommand{\old}[1]{{\oldcolor\sout{#1}}}

\ifthreedvfinal\fi
\begin{document}

\title{3D Neighborhood Convolution: \\ Learning Depth-Aware Features for RGB-D and RGB Semantic Segmentation}

\author[$\dag$]{Yunlu Chen}
\author[$\ddag$]{Thomas Mensink}
\author[$\dag$]{Efstratios Gavves}
\affil[$\dag$]{University of Amsterdam}
\affil[$\ddag$]{Google Research, Amsterdam}
\affil[ ]{\tt\small {\{y.chen3, e.gavves\}@uva.nl, mensink@google.com}}


\maketitle

\begin{abstract}
A key challenge for RGB-D segmentation is how to effectively incorporate 3D geometric information from the depth channel into 2D appearance features.
We propose to model the effective receptive field of 2D convolution based on the scale and locality from the 3D neighborhood.
Standard convolutions are local in the image space ($u, v$), often with a fixed receptive field of 3x3 pixels.
We propose to define convolutions local with respect to the corresponding point in the 3D real world space ($x, y, z$), where the depth channel is used to adapt the receptive field of the convolution, which yields the resulting filters invariant to scale and focusing on the certain range of depth.

We introduce 3D Neighborhood Convolution (3DN-Conv), a convolutional operator around 3D neighborhoods. 
Further, we 
can use estimated depth to use our RGB-D based semantic segmentation model from RGB input.
Experimental results
validate that our proposed 3DN-Conv operator improves semantic segmentation, using either ground-truth depth (RGB-D) or estimated depth (RGB).
\end{abstract}

\section{Introduction}


Most deep networks specialized on semantic segmentation currently follow a fully convolutional architecture \cite{long2015fully,badrinarayanan2017segnet, lin2017refinenet, chen2017deeplab,zhao2017psp, chen2017deeplabv3} on 2D images and return pixel-level classifications. However, as shown in \cite{eigen2015predicting, long2015fully, hazirbas2016fusenet}, semantic segmentation improves when 3D or depth information is available by specialized hardware. One way is that the local segmentation boundary can be refined by the scene geometry, where there is an occlusion boundary in the 2D image projection. 
More generally, the global high-level semantics can benefit from the 3D scene distribution by removal of the possible projection ambiguity. In this work, we investigate if and how we can embed 3D scene information into the 2D convolution in RGB images.

\begin{figure}
    \centering
    \includegraphics[width=1\linewidth]{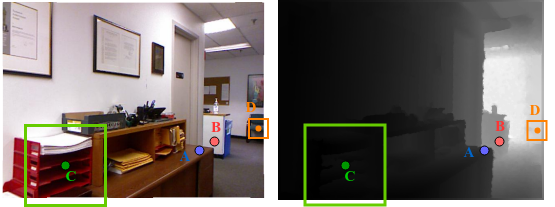}
    \caption{The concept of 3D neighborhood and its difference from 2D neighborhood. \emph{Locality from depth}: {\it A} and {\it B} are neighbours in 2D image but not in 3D space; \emph{Scale from depth}: {\it D} is further away than {\it C}, so the 3D neighborhood of {\it D} is smaller on 2D image than {\it C}. For both cases we can find a explicit cue from the depth value.
    }
    \label{fig:figure1}
\end{figure}


Recently, great progress has been witnessed in deep learning on 3D data, such as voxels~\cite{song2017ssc,chang2015shapenet} and point clouds~\cite{qi2017pointnet, qi2017pointnet++}. Yet, 3D data face problems which prevent it from large-scale or real-time usage.
This changes our attention to 2.5D representations in the form of depth maps and RGB-D data for two reasons.
On the one hand, processing 2.5D data is almost as computationally efficient as 2D data, contrast to 3D representations that blow up computations.
On the other hand, 2.5D RGB-D data can easily be acquired by either low-cost 2.5D commercial depth sensors like Kinect or disparity maps from binocular cameras,
What is more, in the absence of sensory depth, monocular depth estimation methods~\cite{eigen2014depth,eigen2015predicting,laina2016deeper,xu2017mscrf,fu2018dorn} has recently been able to provide reasonable depth maps, even with RGB images alone.
The better availability and efficiency of 2.5D data render them an inexpensive and yet effective solution for incorporating geometry information.

\begin{figure*}[t!]
    \centering
    \includegraphics[width=0.85\linewidth]{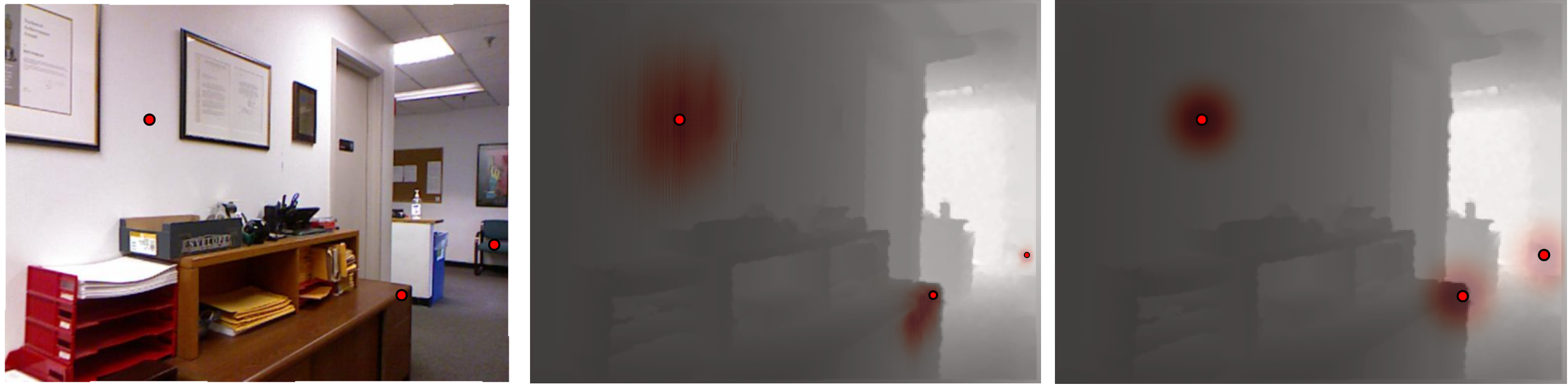}
    \caption{Visualization of the effective receptive fields \cite{luo2016erf} sampled at three certain points. Left: original image. Middle: effective receptive field of 3DN-Conv. Right: effective receptive field of standard convolution. Results in six sequential layers of 3DN-Convs.}
    \label{fig:erf}
\end{figure*}

A reasonable question, therefore, is how to effectively incorporate depth into a model, such that to learn depth-aware features amenable for scene semantic segmentation.
In this paper, we integrate depth into the 2D convolutional operation.
We do not just add depth as additional input channels as usual, but we define a \emph{3D neighborhood} and modify the 2D convolutional filters with it. 
As illustrated in Figure~\ref{fig:figure1}, the two properties of 3D neighborhood are  \emph{depth locality} and \emph{scale}. The depth locality comes from the locality in the 3D space; the varying scale in 2D image plane is due to the requirement of scale consistency in 3D. 
Depth locality and scale are important because they determine the receptive field.
Taking the perspective that convolution operation is a \emph{local function}, we manage to learn more 3D-aware features by tuning the receptive field of the 2D convolution according to the 3D neighborhood.

We present 3D Neighborhood convolutions (3DN-Conv)
In 3DN-Conv the receptive fields dynamically adapt to the local depth structure.
To be specific, the size of the receptive field is inversely proportional to the depth at the same pixel location, determined by the rule of \emph{scale},
and only the region within a certain range of depth are incorporated to the convolution, according to the rule of \emph{depth locality}.
The visualization in Figure~\ref{fig:erf} supports that the effective receptive field of our 3DN-Conv bears awareness of both properties of scale and depth locality.
We also use the depth map returned by the proposed Depth Discriminative Feature Network (D-DFN) for semantic segmentation, to obtain depth-aware pixel-level label predictions.

We make the following contributions.
\emph{First}, we propose 3D Neighborhood Convolutions (3DN-Conv) a novel spatially variant and depth-aware convolutional operator.
The proposed convolution considers depth as a cue for both locality along the camera z-axis and the receptive field scale of the kernel.
To the best of our knowledge, we are the first to propose a convolution operation that explicitly considers both aspects.
Experiments show that our 3DN-Conv is more effective than other methods to incorporate depth for semantic segmentation.
\emph{Second}, we propose a depth estimation model D-DFN that recovers accurate local depth gradients and sharp 3D edges, compared with the state-of-the-art depth estimation algorithms.
\emph{Third}, we show that the proposed depth estimation can be successfully combined with RGB-D segmentation models and reach accuracies for which normally one requires depth information by specialized hardware sensors.

\section{Related Works}


\vspace{4pt}
\paragraph{RGB-D semantic segmentation}



RGB-D segmentation extends RGB semantic segmentation~\cite{long2015fully,badrinarayanan2017segnet, chen2017deeplab,zhao2017psp,chen2017deeplabv3,yu2018dfn}. 
A widely applied method (e.g. in \cite{long2015fully, eigen2015predicting, chu2018surfconv}) is to 
%
%
encode depth into hand-craft HHA features \cite{gupta2014hha}: horizontal disparity, height above ground and angle with gravity. 
Hazirbas~\etal~\cite{hazirbas2016fusenet} uses a separate encoder to process depth channel and fuses the feature maps at every stage. Qi~\etal~\cite{qi20173d}
built a 3D k-nearest neighbor graph network on point clouds with extracted features from a CNN.

In this work, we focus on using depth to build spatially variant convolutional filters, with the advantage to learn better geometry-aware features by modeling the receptive field of the 2D convolution in accordance with the 3D local neighborhood. Following this trend, Wang and Neumann~\cite{wang2018depthaware} augments the standard convolution by adding depth similarity as a weight term to consider the locality on dimension of depth. Different from \cite{wang2018depthaware}, depth can also be a cue for scale~\cite{porzi2017depth, chu2018surfconv}: pixels with larger depth values are processed with convolutions with a smaller receptive field. 

Unlike those works, our 3D neighborhood is the first concept framework to explicitly cover both the scale \emph{and} the locality from depth in theory, which learns better 3D-aware features, as we will show experimentally.

%
%

\paragraph{Supervised monocular depth estimation}
We denote with 
monocular depth estimation (MDE) the task to recover the depth map from a single RGB image as input. While handcrafted features and probabilistic graphical models are used for MDE in early years~\cite{saxena2006learning,saxena2008make3d}, recent methods benefit from learned structural deep features for local and global contextual information. Eigen \etal\cite{eigen2014depth,eigen2015predicting} proposed a coarse-to-fine network to refine the output depth from a coarse prediction stage by stage.
Laina ~\etal\cite{laina2016deeper} introduced residual block design into fully convolutional network. Xu ~\etal\cite{xu2017mscrf} fuses multi-scale depth output in different stages by Conditional Random Fields (CRFs).
Fu ~\etal \cite{fu2018dorn} proposed ordinal regression as multiple binary classifications for pixel-wise depth, which achieves current state-of-the-art performance.



\paragraph{Adaptive Convolutions}
Our work also relates to generic adaptive convolutions, where local image filters are defined based on a set of different basis functions~\cite{jacobsen16cvpr}, features of the previous layer~\cite{verma18cvpr,brabandere16nips}, the correlation of the input or features~\cite{wang2018nonlocal, Shevlev_2019_cooccur}, or a different modality~\cite{wang2018depthaware}. In particular, the most similar variants to our methods are the spatial sampling location methods, where filters are locally adaptive based on the spatial structure of the data itself~\cite{dai17iccv, zhu2019deformablev2, zhang2017scale, recasens2018zoom}.
In contrast to these works that learn the local filter from the data, our convolution filter is defined using depth as privileged information.

\section{3D Neighborhood Convolution}
In this section, we describe our 3D Neighborhood Convolution (3DN-Conv) models.

\subsection{3D neighborhood in RGB-D coordinates}

Every 2D pixel location on the image frame corresponds to local neighborhood in the 3D space.
While a convolutional filter normally has a predefined receptive field, 
we argue that the spatial extent of the convolutional filter should depend on the 3D neighborhood around the real-world point projected into a particular pixel.
To define the 3D neighborhood around an image location, we consider the 3D cube around real-world point $\mathbf{p} = (p_x, p_y, p_z)$, with a radius of $\sigma$.
Which we subsequently approximate from 2D image coordinates and depth $(p_u, p_v, d)$, resulting in depth aware 2D filters in image space.

Under the constraints of a classic pin-hole camera model, the image coordinates of real-world point $\mathbf{p}$ are:
 \begin{equation}
 \left\{ 
 \begin{array}{rcl}
 p_u & = &\mu\dfrac{p_x}{p_z},\vspace{6pt}\\
  p_v & = &\mu\dfrac{p_y}{p_z},\\
   d & = & p_z,
 \end{array}
\right.
      \label{eq:projection2d}
 \end{equation}
where $\mu$ relates to the camera focal length.
A $\sigma$ neighborhood in 3D around $\mathbf{p}$ can be approximated by the following 2D image neighborhood: 
 \begin{equation}
    \left\{ 
\begin{array}{l}
p'_u \in [\mu\frac{p_x - \sigma}{p_z},  \mu\frac{p_x + \sigma}{p_z}]=[p_u - \mu \frac{\sigma }{d}, p_u + \mu \frac{\sigma }{d}], \\[2mm]
p'_v \in [\mu\frac{p_y - \sigma}{p_z}, \mu\frac{p_y + \sigma}{p_z}]=[p_v - \mu \frac{\sigma }{d}, p_v + \mu \frac{\sigma }{d}].
\end{array}
\right.
\label{eq:projection2dneighborhood}
\end{equation}
Which shows that a 3D cube with size $\sigma$ corresponds to an image based convolution operator with a receptive field of:
 \begin{equation}
 \vspace{-8pt}
    \Delta p_u = \Delta p_v = \mu\dfrac{\sigma}{d},
    \label{eq:receptivefield}
 \end{equation}
which suggest that the receptive field of the 2D kernel should be inversely proportional to the depth $d$,
instead of $\Delta p_u = \Delta p_v$ being a predefined filter size of the convolution.
The required depth value $d$ can be estimated from the z-buffer of the RGB-D image.

The value of depth $d$ is not influenced by camera projection, yet a direct measurement of the 3D world, therefore: 
\vspace{-2pt}
\begin{equation}
\Delta_{d} = \sigma \\ \textrm{, and } 
d' \in [d -\sigma,\quad  d + \sigma].
\vspace{-1pt}
\label{eq:2}
\end{equation}
This shows that a local neighborhood on the depth channel is defined by the radius of the 3D neighborhood. 

In the rest of this section we use the insights from Eq.~\ref{eq:receptivefield} and Eq.\ref{eq:2} for the design of our 3DN-Conv operator.

\subsection{Depth locality}
\label{sec:depth_locality}

For clarity of presentation, we first formulate the standard convolutional filter for pixel $i$ as follows:
   \vspace{-4pt}
\begin{equation}
    \mathbf{y}_i = f \left(\mathbf{b} + \sum_{j \in \mathcal{N}_i} \mathbf{W}_{n_{ji}}  \ \mathbf{x}_j \right),
    \vspace{-2pt}
\end{equation}
where $\mathbf{y}_i$ denotes the $D$ dimensional output vector for pixel $i$, and $\mathbf{x}_j$ denotes the $C$ dimensional input vector for pixel $j$. Furthermore, $f(\cdot)$ denotes the activation function (\eg ReLU), $\mathbf{b}$ denotes the bias ($\mathbf{b} \in \mathbb{R}^{1 \times D}$), $j \in \mathcal{N}_i$ enumerates over the spatial neighborhood around pixel $i$, $n_{ji}$ denotes the relative position of $j$ with regard to $i$ inside the neighborhood to select the relevant slice of the filter $\mathbf{W}$, where for each relative location we learn a filter $\mathbf{W} \in \mathbb{R}^{D \times C}$.

\vspace{-8pt}
\paragraph{Modeling depth locality}
From Eq.~\ref{eq:2} we know that the locality of depth should be within a range of $[d_i - \sigma, d_i + \sigma]$. 
To use this as \emph{effective} filter size, we use a function that decays with depth as the local window to reweigh the local convolution kernel, which is equivalent to reweighing the local input features. 

\begin{figure}
    \centering
    \includegraphics[width = 0.75\linewidth]{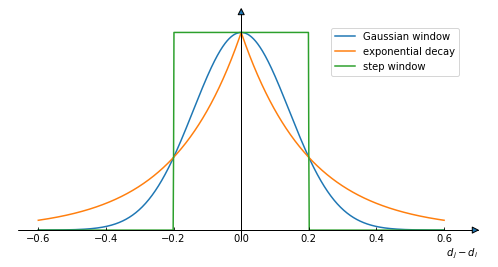}
    \caption{Illustrating different locality functions. A sharp step function (green) loses flexibility to incorporate non-local features. The exponential decay function (orange), used in \cite{wang2018depthaware}, has too long tails. We use the Gaussian curve (blue), since it is in accordance with the effective receptive field theory~\cite{luo2016erf}.}
    \vspace{-8pt}
    \label{fig:window}
\end{figure}

We evaluate different functions for reweighing, see Figure~\ref{fig:window}. Intuitively a step function, which uses a hard threshold on the locality seems an obvious choice, yet this fails in practice. We argue this is because 2D receptive field can gradually grow along with more convolution kernels, while receptive field on depth can hardly grow with 2D convolution. 
By noting that the effective receptive field follows the shape of a Gaussian distribution, see ~\eg~\cite{luo2016erf}, we choose to model the window function by a 1-D Gaussian function $N(d_i,\sigma)$, where $\sigma$, the size of the expected 3D neighborhood, is used as the standard deviation, the size of the Gaussian window.
This yields the following convolutional filter:
\begin{align}
\vspace{-8pt}
    \mathbf{y}_i &= f\left(\mathbf{b} + \sum_{j \in \mathcal{N}_i} \mathcal{L}_{ji} \mathbf{W}_{n_{ji}} \ \mathbf{x}_j \right), \ \textrm{where}
    \label{eq:locality_filter}
    \\
    & \mathcal{L}_{ji} = {\rm exp}(\dfrac{d_j - d_i}{\sigma})^2.
    \label{eq:locality_weight}
\end{align}
In the following we denote this local convolutional filter with: $\mathbf{W}_{n_{ji}}^{\mathcal{L}} = \mathcal{L}_{ji} \mathbf{W}_{n_{ji}}$.

In standard 2D CNNs the effective receptive field of a kernel grows per layer, due to the aggregation of information and the use of pooling layers. 
Therefore, we scale $\sigma$ per layer with respect to the size of 2D convolution kernel, so that the effective 3D neighborhood remains similar. According to Eq.~\ref{eq:receptivefield}, the size of the 2D kernel is proportional to the desired size of the 3D neighborhood. However, in different layers in the ConvNet, as the features are usually downsampled in 2D spatial resolution stage by stage, the relative size of the canonical 2D kernel, which we regard as the size of the 3D neighborhood $\sigma$, is actually enlarged as the network goes deeper. So $\sigma$ varies with layer in the network. The exact scaling factor depends on the used architecture and is clarified in the supplementary material.

\begin{figure*}[htbp]
    \centering
    \includegraphics[width=0.85\linewidth]{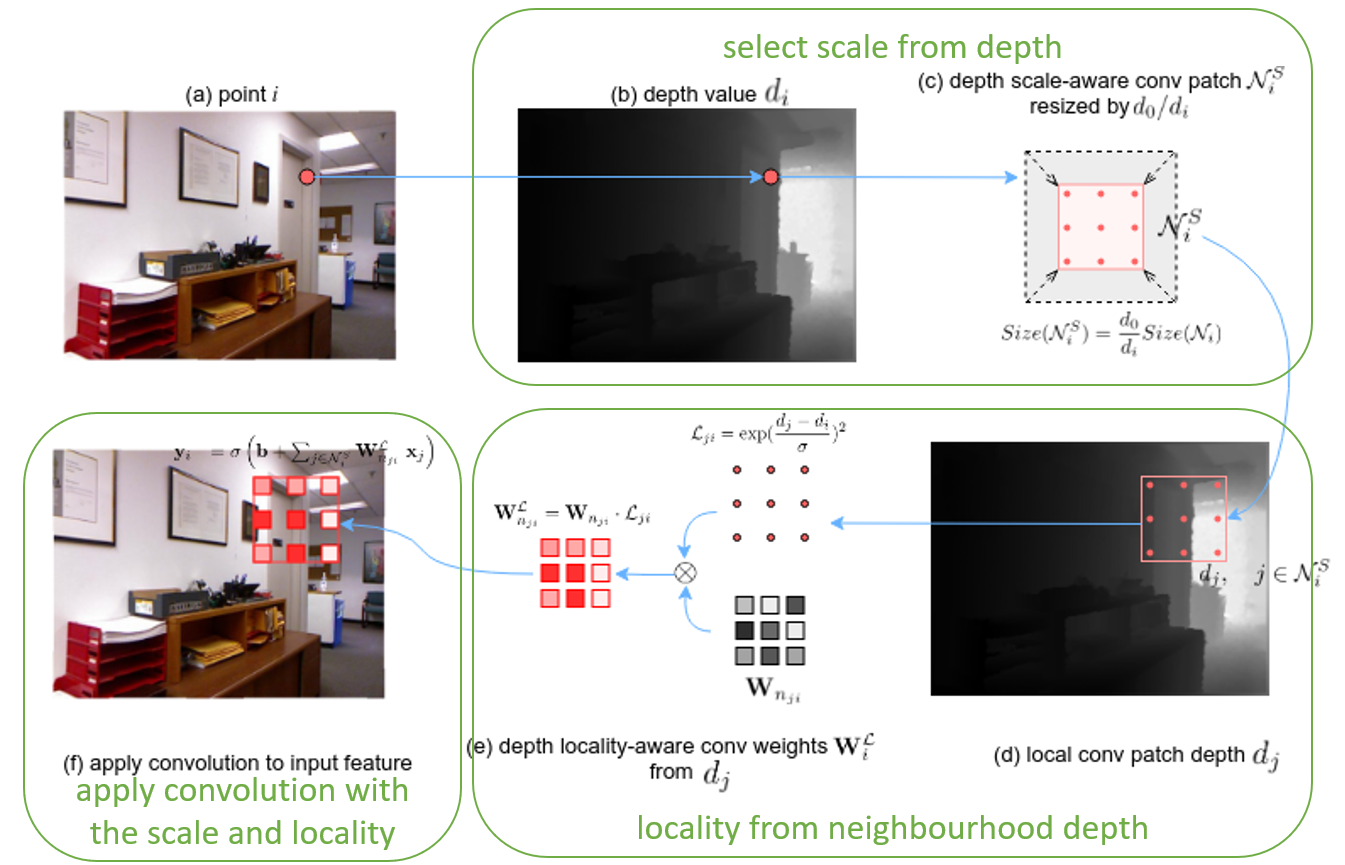}
    \caption{A practical overview of 3DN-Conv. The depth \emph{scale}-aware conv patch $\mathcal{N}_i^S$ and the depth \emph{locality}-aware conv weights $\mathbf{W}_i^\mathcal{L}$ (step (c) and step (e)) are calculated
    from depth (step (b),  by Eq.~\ref{eq:scaled_neighborhood} and Eq.~\ref{eq:scaled_filter_size}, and step (d), by Eq.~\ref{eq:locality_filter} and Eq.~\ref{eq:locality_weight}).
    }
    \vspace{-5pt}
    \label{fig:pipe}
\end{figure*}

\subsection{Neighborhood scale selection}
We know that the receptive field of the 2D convolution $\Delta_d$ should be inversely proportional to the local depth value $d$, see Eq.~\ref{eq:receptivefield}. 
In order to incorporate this into our design, we choose to bilinearly resample the convolutional patch, \ie, the 2D local neighborhood $\mathcal{N}_i$ to a rescaled version $\mathcal{N}_i^{S}$.
This yields the following convolutional filter:
\vspace{-8pt}
\begin{equation}
    \mathbf{y}_i = f\left(\mathbf{b}_{n_{ji}} + \sum_{j \in \mathcal{N}^{S}_i} \mathbf{W}_{n_{ji}}^{\mathcal{L}}  \ \mathbf{x}_j \right),
    \label{eq:scale_filter}
\end{equation}
\vspace{-6pt}
where $\mathcal{N}_i^{S}$ is the scaled 2D neighborhood, defined by:
\begin{align}
        \mathcal{N}_i^{S}:= &\{j | \delta (i,j)\leq r^S \}, \ \ \textrm{where}
        \label{eq:scaled_neighborhood}
        \\
        & r^S = \dfrac{d_0}{d_i} r_0
        \label{eq:scaled_filter_size}
\end{align}
in which $\delta(i,j)$ denotes the distance of the two points $i$ and $j$ in 2D space. $r^S$ is the scaled kernel size, and $r_0$ is the original kernel size and it usually equals to dilation rate. In practice, we need to set a canonical depth value $d_0$ as a hyperparameter so that the  size of the local convolution patch adapts according to Eq.~\ref{eq:scaled_filter_size}.


We include an overview in Figure~\ref{fig:pipe} on how we build our 3DN-Conv to model the scale and depth locality. In practice, we first scale the convolution patch size and sample the scaled neighborhood by Eq.~\ref{eq:scaled_neighborhood} and Eq.~\ref{eq:scaled_filter_size}, and then adapt the local filter to incorporate depth locality by Eq.~\ref{eq:locality_filter} and Eq.~\ref{eq:locality_weight}.

\subsection{Leverage RGB edges}
The first few layers of standard convolutional networks extract edge features and other low level features. 
While our 3D-Neighborhood Convolution incorporates features in accordance with their real world spatial location, the limitation of looking at only 3D neighbors could be to lose edge kind features. Especially the edges caused by depth occlusions, since both sides of the occlusion edge are not regarded as the neighborhood in our convolution being aware of 3D neighborhood. 

In order to leverage edge and other low-level features, we combine standard 2D convolution filters, denoted by $\mathbf{W}^\mathcal{E}$ for edge features with the proposed 3DN-Conv:
\vspace{-8pt}
\begin{align}
    \mathbf{y}_i &= f\left(\mathbf{b} + \sum_{j \in \mathcal{N}_i^{S}} \mathbf{W}_{n_{ji}}^{\mathcal{L}}  \ \mathbf{x}_j +  
    \mathbf{W}_{n_{ji}}^{\mathcal{E}}  \ \mathbf{x}_j \right).
    \label{eq:wedge}
\end{align}
Note that we deliberately use a separate set of parameters for $ \mathbf{W}^{\mathcal{E}}$ without any weight sharing with $ \mathbf{W}^{\mathcal{L}}$ to ensure the two kernels learning discriminate features.
This joint convolution is only applied in the first stage of convolution layers \verb'conv_1' in the ResNet backbone.

\vspace{-5pt}
\paragraph{In depth comparison with \cite{wang2018depthaware}}
Our proposed 3DN-Conv bears resemblance to the depth-aware convolution operator proposed by Wang and Neumann \cite{wang2018depthaware}. The depth-aware spatial-variant convolution reweighs features to be convolved by an exponential function that measures the difference in depth.
We clarify some significant differences:
(i) We explicitly model a scaling function for our kernel, while \cite{wang2018depthaware} only considers depth similarity. 
This is the largest difference and traces back to the fact that our motivation being different from \cite{wang2018depthaware} as we design our kernel based on modeling the 3D neighborhood. 
(ii) Our local weight function is a Gaussian depth locality function to resemble the shape of the effective receptive field. While \cite{wang2018depthaware} uses an exponential decay function (see Figure~\ref{fig:window}). 
(iii) We use varying depth locality window size $\sigma$ in different layers in the network while in \cite{wang2018depthaware} the window size is fixed in the network (see the last paragraph in Section~\ref{sec:depth_locality}). 
(iv) We include the edge convolution component to enhance low-level features.

\section{Learning Depth for RGB-D Segmentation}


In RGB-D segmentation, depth provides extra information such as 3D boundaries to eliminate some ambiguities in projected 2D images.
Unfortunately, it is not always possible to have sensory depth as this requires specialized equipment.
For this reason, we propose to first estimate depth, followed by semantic segmentation using our 3DN-Conv described in the previous section.
The estimated depth map should be locally correct, including fine local details, sharp depth boundaries and consistent depth gradients, in order to guide the semantic segmentation model. 

\begin{figure}
    \centering
    \includegraphics[width = 1.05 \linewidth]{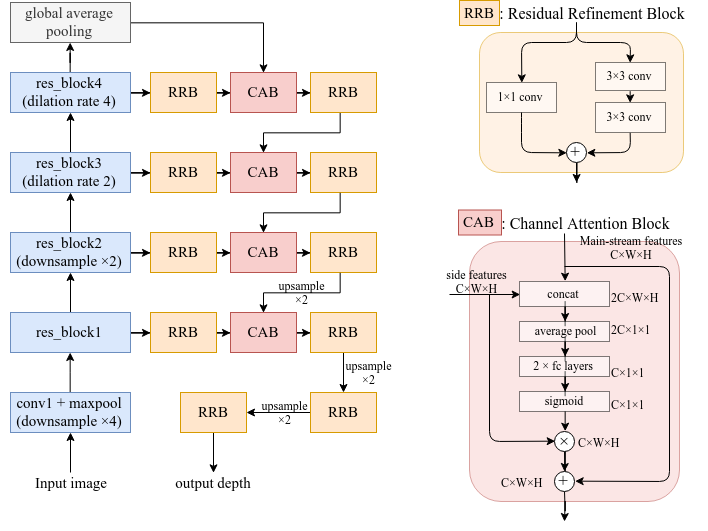}
    \caption{Illustration of Depth Discriminative Feature Network (D-DFN) for depth prediction. We experimentally show that estimating depth from RGB and then use our depth-aware 3DN-Conv for semantic segmentation improves over using RGB segmentation.
    }\vspace{-10pt}
    \label{fig:ddfn}
\end{figure}

\subsection{Depth Discriminative Feature Network}

The task of depth estimation shares some common properties with semantic segmentation.
For one, both are pixel-wise prediction tasks.
Also, both tasks require not only recognizing both global contextual information as well as local details, but also a successful fusion strategy for combining the two.
It stands to reason, therefore, that depth estimation and semantic segmentation may can benefit from each other when it comes to designing a suitable network architecture.

Inspired by the recently proposed semantic segmentation network Discriminative Feature Network (DFN) \cite{yu2018dfn}, we describe our proposed a novel model for estimating depth specialized for encoding local depth structure. 
In DFN architecture, the Channel Attention Block module (CAB)  is proposed to incorporate features at multiple scales, with the idea originate from channel attention \cite{hu2018senet}. 




We describe the architecture of our D-DFN. 
The model structure is illustrated in Figure~\ref{fig:ddfn}, comprising an encoder and a decoder architecture.
The encoder is a ResNet-101 backbone, followed by a global pooling layer for capturing global information. The Residual Refinement Block upconvolves the high-level features to the output stage by stage. The Channel Attention Block incorporates features from different stages with channel-attention.
See supplementary material for a more detailed comparison of our D-DFN and the original DFN architecture.


\vspace{-8pt}
\paragraph{Training Loss.}
We adopt the $L_1$ loss for depth error 
\begin{equation}
\vspace{-5pt}
 L_{depth} = \frac{1}{N} \sum_i {|d_i - \hat{d_i}|}.
 \vspace{-5pt}
\end{equation}
We also consider an auxiliary loss for incorporating depth gradients and thus learning fine local depth details.
The auxiliary loss takes the form 
\begin{equation}
    L_{grad} = |\nabla_x d_i - \nabla_x\hat{d_i}|+ |\nabla_y d_i - \nabla_y\hat{d_i}|,
\end{equation}
where $N$ denotes the total number of pixels, $d_i$ the predicted depth value at the $i$-th pixel and $\hat{d_i}$ is the ground truth.
This auxiliary loss enforces smoothness, by penalizing inconsistent depth gradients~\cite{li2017twostream, hu2019revisiting}.
The final loss is a weighted combination: $L_{depth} + \lambda_{grad} L_{grad}$, with $ \lambda_{grad} = 1$.

\section{Experimental Results}

We provide an extensive evaluation of our 3DN-Conv.
We adopt Deeplab V3~\cite{chen2017deeplabv3} and DFN~\cite{yu2018dfn} as our baseline semantic segmentation models. The popular Deeplab V3 is used for most of the experiments and ablation studies, while DFN is for compare with state-of-the-art methods as it is empirically more powerful for complex scene segmentation. 
Following ~\cite{long2015fully}, we adopt the following common metrics to evaluate semantic segmentation: 
mean intersection-over-union (mIoU), mean accuracy (mAcc) and pixel accuracy (Acc). 
For more implementation details we kindly ask readers to refer to the supplementary material.
\vspace{5pt}

\noindent{\bf Datasets}
Evaluation is performed on two RGB-D segmentation datasets:
NYUDv2 \cite{silberman2012nyu} and KITTI \cite{xu2016kitti11}.
NYUDv2 contains a total of 1,449 RGB-D image pairs from 464 different scenes. The dataset is divided into 795 images from 249 scenes for training and 654 images from 215 scenes for testing. We use the 40-class segmentation setting \cite{gupta2013seg40}.
KITTI  provides parallel camera and LIDAR data for outdoor driving scenes. We use the semantic segmentation annotation proposed in \cite{xu2016kitti11}, which contains 70 training
and 37 testing images and annotations in 11 classes.

\subsection{Depth-aware convolutions: ablation study}

\begin{table}[]
\begin{center}
\begin{tabular}{lccc}

\toprule[1.5pt]

                                                   & mIoU(\%) & mAcc(\%) \\ \hline
Baseline (RGB only)                                & 29.6     & 41.2       \\ 
Depth-aware conv \cite{wang2018depthaware}         & 36.9     & 49.0       \\ \hline
 $\mathbf{W}^{\mathcal{L}}$             & 37.2     & 49.4        \\
 $\mathbf{W}^{\mathcal{L}}$+ $\mathcal{N}^{\mathcal{S}}$ {\small (discretized dilation rate)} & 38.5     & 51.2         \\
 $\mathbf{W}^{\mathcal{L}}$ + $\mathcal{N}^{\mathcal{S}}$ {\small(bilinear sampling)}     & 39.1     & 51.7     \\
3DN-Conv (final)     & {\bf 39.3}     & {\bf 52.4}    \\
\bottomrule[1.5pt]

\end{tabular}
\end{center}
\caption{Semantic segmentation performance on NYUDv2, with components in 3DN-Conv, trained from scratch.}
\vspace{-10pt}
\label{tab:ablation}
\end{table}

We examine the importance of each of the proposed components of 3DN-Conv.
We use our implementation of depth-aware conv in~\cite{wang2018depthaware} and applied on the same ResNet50-DeeplabV3 model.
We evaluate the performance of semantic segmentation on the NYUDv2 dataset.

The results are shown in Table~\ref{tab:ablation}. 
We observe the following:
First both depth-aware convolution methods improve over the RGB segmentation baseline by a large margin, as was also noted in~\cite{wang2018depthaware}.
Our improved 3DN-Conv, using $\mathbf{W}^{\mathcal{L}}$ with a Gaussian window brings an additional improvement over the exponential decay in \cite{wang2018depthaware}.
Furthermore, we examine integrating scale adaptation to the proposed depth-aware convolution in two ways.
First, the integration of scale can be done in a discretized manner.
The depth intervals are first binned.
Then, different dilation rate kernels are considered, similar to~\cite{chu2018surfconv}.
The second way of integrating scale is by considering bilinear rescaling of the convolutional receptive field, as described in Sec. 3.2.
With the bilinear rescaling a continuous value for the scale is returned, thus allowing for more fine-tuned convolutions.
In any case, per location a single scale is selected.
We observe that adding scale adaptation improves the depth-aware convolutions considerably, especially when considering continuous scale values.
This is not surprising, as the distribution of depth values is highly non-uniform, thus it is not straightforward how to bin fairly.
Last, when considering also the RGB-only component for the convolution, $\mathbf{W}^{\mathcal{E}}$, the performance improves further in all three metrics.

We further evaluate on KITTI dataset~\cite{geiger2013kitti, xu2016kitti11} 
to examine our method in various types of scenes. As in Table \ref{tab:kitti}, we see that our 3DN-Conv outperforms both RGB baseline and the depth-aware convolution in \cite{wang2018depthaware}.

We conclude that the proposed depth-aware convolutions improve semantic segmentation, especially when considering adaptive scaling of the receptive field, as well as additional RGB-only filters.

\begin{table}[]
\begin{center}
\begin{tabular}{lccc}

\toprule[1.5pt]

                                                   & mIoU(\%) & mAcc(\%) \\ \hline
Baseline (RGB only)                                & 35.1     & 44.1         \\ 
Depth-aware conv \cite{wang2018depthaware}         & 38.1     & 48.0        \\ 
3DN-Conv (this paper)  & {\bf 39.2}     & {\bf 50.8}    \\
\bottomrule[1.5pt]

\end{tabular}
\end{center}
\caption{Semantic segmentation on KITTI, trained from scratch}
\vspace{-10pt}
\label{tab:kitti}
\end{table}

\subsection{Fusing depth \& RGB for segmentation}

Next, we are explore the optimal way of integrating depth information for semantic segmentation.
Specifically, we consider the following choices: (i) \textbf{early-fusion} of HHA features (ii) \textbf{late-fusion} of HHA features, (iii) \textbf{feature reweighting (modulation)}: feature map rescaled by depth, with a simple linear model, and (iv) \textbf{feature reweighting (non-local)}: feature map rescaled by depth, with non-local attention~\cite{wang2018nonlocal}. See supplementary material for the details.

For all the aforementioned methods we use the same backbone architecture and training pipeline.
We compare the four methods of fusion above with the depth-aware convolutions from \cite{wang2018depthaware}, as well as our 3D-Neighborhood convolution. 
We report results in Table~\ref{tab:incorporate}.

We observe that the methods which perform convolution-level incorporation of depth, including the depth-aware convolution~\cite{wang2018depthaware} and our 3DN-Conv, outperforms other methods by a large margin.
The other methods help to improve the performance compared to the baseline by introduction of depth information.
However, the fusion methods are simply processing the HHA encoding of depth as extra channels in the network, and 
the assumptions of the feature reweighting method (see supplementary material) are too simple to capture the influence from the depth map. Overall, these methods to incorporate depth into deep network are regardless of the essence of geometry.
We conclude that our 3DN-Conv is able to capture better depth-aware features by convolution-level incorporation and an explicit modeling of 3D geometry.

\begin{table}[]
\begin{center}
\begin{tabular}{lccc}
\toprule[1.5pt]
                                                   & mIoU(\%) & mAcc(\%) \\ 
                                                   \hline
Baseline (RGB only)                                & 29.6     & 41.2       \\ \hline
HHA (early-fusion)                                       & 32.0     & 43.5       \\
HHA (late-fusion)                                        & 33.7     & 44.4       \\
Feature reweighting (modulation)                   & 31.5     & 43.4       \\
Feature reweighting (non-local)                  & 32.6     & 47.2     \\
Depth-aware conv  \cite{wang2018depthaware}        & 36.9     & 49.0     \\ \hline
3DN-Conv (this paper)                                             & {\bf 39.3}     & {\bf 52.4}     \\
\bottomrule[1.5pt]

\end{tabular}
\end{center}
\caption{Semantic segmentation performance on NYUDv2, with different method to incorporate depth into RGB network, on NYUDv2, trained from scratch.  The spatial-variant kernel methods (\cite{wang2018depthaware} and ours) outperform other methods by a large margin. }
\vspace{-10pt}
\label{tab:incorporate}
\end{table}

\subsection{Segmentation with estimated depth}

Next, we evaluate whether estimated depth can be used to improve semantic segmentation, in a similar way like depth returned by specialized sensory equipment. 
First, we examine the quality of the depth estimations.
Then, we examine the benefits of using depth to semantic segmentation.

\vspace{-8pt}
\paragraph{Depth estimation}
Following~\cite{eigen2014depth}, we evaluate the depth estimation performance by taking the $304 \times 228$ center crop out of the downsampled image.
As the goal is to use depth as a cue for semantic segmentation, it is important to have accurate local depth estimation.
To this end, we evaluate depth estimation with global and local metrics, that is (i) Root mean squared error (rms): $\sqrt{\frac{1}{N} \sum_i (d_i - \hat{d_i})^2}$, (ii) Mean log10 error (log): $\frac{1}{N} \sum_i {||{\rm log}_{10}d_i - {\rm log}_{10}\hat{d_i}||}$. (iii) Gradient error (grad) $|\nabla_x d_i - \nabla_x\hat{d_i}|+ |\nabla_y d_i - \nabla_y\hat{d_i}|$.
Whereas the root mean squared error and the mean log10 error focus on the global evaluation of depth estimation, the local gradient error measures how well the local depth details are predicted.
We report results in Table \ref{tab:depth}.

\begin{figure}
    \centering
    \includegraphics[width=1.0\linewidth]{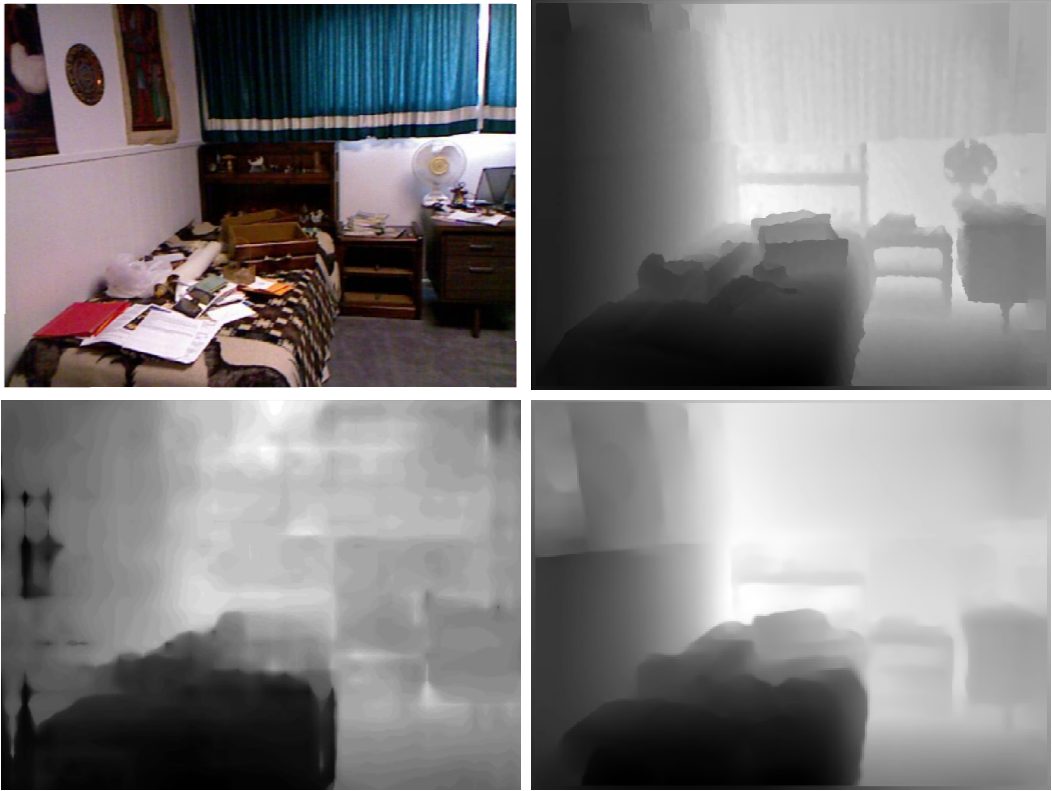}
    \caption{Depth estimation examples. Left top: input image. Right top: ground truth. Left bottom: predicted depth by DORN \cite{fu2018dorn}. Right bottom: prediction by our D-DFN. Prediction by DORN shows artifacts and the local depth edges or gradient consistency are not well recovered, while our D-DFN output much sharper local depth structure.}
    \label{fig:depth}
\end{figure}

We observe that the proposed D-DFN improves in terms of both the global mean log10 error, however, it performs slightly worse in terms of root mean squared error. 
We attribute this to the fact that the logarithm scale normalizes the possible output values in a more reasonable range, in which regression is easier.
Further, the proposed method improves the local gradient error by a noticeable 30\% compared to \cite{fu2018dorn}.
This is important, as for semantic segmentation the local depth structure indicates the presence of semantic boundaries and can help with ambiguities.
We corroborate this by showing some qualitative results in Figure \ref{fig:depth}, where the proposed D-DFN returns smoother outputs and finer local depth details.


\begin{table}[]
\begin{center}
\begin{tabular}{@{\extracolsep{4pt}}lccc@{}}
\toprule[1.5pt]
& \multicolumn{2}{c}{Global} &  \multicolumn{1}{c}{Local} \\
\cline{2-3} \cline{4-4}
                 & RMS & log10 & Local grad \\ 
                                                   \hline
Laina \etal \cite{laina2016deeper}        & 0.573&0.055 & 0.137 \\ 
DORN \cite{fu2018dorn}               & {\bf 0.509}&0.051&0.140 \\\hline
D-DFN (this paper)                          & 0.528& {\bf 0.049}     & {\bf 0.092} \\
\bottomrule[1.5pt]

\end{tabular}
\end{center}
\caption{Depth estimation performance.
Our D-DFN model achieves global results that is close-to the state-of-the-art model DORN, with significantly overwhelming local gradient error.
}
\vspace{-5pt}
\label{tab:depth}
\end{table}

\begin{figure*}[htbp]
    \centering
    \includegraphics[width = 0.96\linewidth]{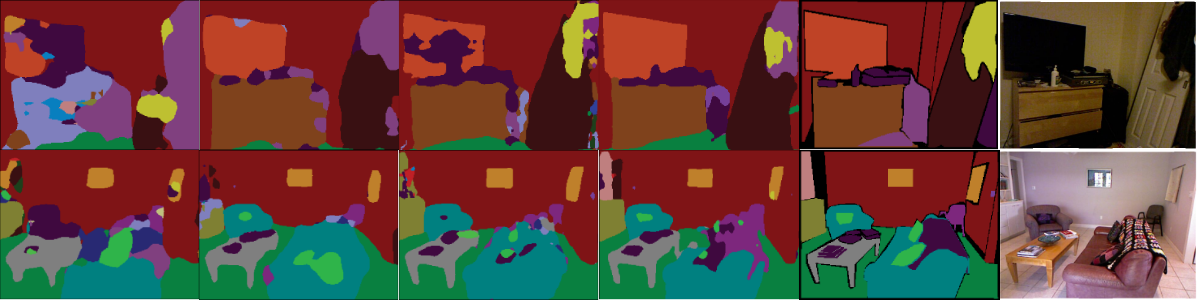}
    \caption{Qualitative results. Columns from left to right: (i) Deeplab V3; (ii) Deeplab V3 + 3DN-Conv; (iii) DFN; (iv) DFN + 3DN-Conv; (v) GT; (vi) image. Deeplab V3 trained from scratch and DFN pretrained on ImageNet. 
    }
    \vspace{2pt}
    \label{fig:qualitative}
\end{figure*}

\begin{table}[]
\begin{center}
\begin{tabular}{lccc}
\toprule[1.5pt]
                                           &  Depth      & mIoU(\%) & mAcc(\%)  \\ 
                                                   \hline
Baseline & /                               & 29.6     & 41.2        \\ \hline
3DN-Conv & Ground Truth                            & 39.3     & 52.4     \\\hline
3DN-Conv  & DORN \cite{fu2018dorn}     & 33.2     & 44.3     \\
3DN-Conv  & D-DFN                           & {\bf 36.4}    & {\bf 48.1}     \\
\bottomrule[1.5pt]

\end{tabular}
\end{center}
\caption{RGB-D segmentation results with different source of depth. Segmentation models are trained from scratch.}
\vspace{-8pt}
\label{tab:est_depth}
\end{table}

\begin{table*}[!ht]
\begin{center}
\resizebox{0.76\textwidth}{!}{
\begin{tabular}{lcccc}
\toprule[1.5pt]
 & Depth at inference time & mIoU(\%) & mAcc(\%) & Acc(\%) \\ 
\hline
RefineNet \cite{lin2017refinenet} & / & 46.5     & 58.9     & 73.6 \\
\hline
FCN \cite{long2015fully} (RGB-HHA)  & Ground Truth & 34.0     & 46.1     & 65.4    \\ 
Qi \etal\cite{qi20173d}  & Ground Truth  & 43.1     & 55.7     & / \\
Wang and Neumann \cite{wang2018depthaware} & Ground Truth & {\bf 48.4}     & 61.1    & / \\ \hline

DFN \cite{yu2018dfn}          & /            & 45.5     & 58.8     & 72.7 \\
DFN + 3DN-Conv & Ground Truth   & 48.2     & {\bf 61.2}     & {\bf 74.8} \\ 
DFN + 3DN-Conv & Estimated    & 47.0     & 59.5     & 74.3 \\
\bottomrule[1.5pt]
\end{tabular}
}
\end{center}
\caption{State-of-the-art in RGB-D segmentation with different source of depth. Our 3D neighborhood convolution successfully leverages the estimated depth by the proposed D-DFN depth estimation network to improve RGB-D semantic segmentation, without requiring sensory depth at inference time.
}
\vspace{-5pt}
\label{tab:sota}
\end{table*}

\vspace{-8pt}
\paragraph{Depth-aware semantic segmentation}
Next, we evaluate the performance of using estimated depth for RGB-D segmentation.
We report results in Table \ref{tab:est_depth}.

We make several observations.
First, we confirm the findings that ground truth depth improves semantic segmentation.
When using the proposed D-DFN to estimate depth and help semantic segmentation, we improve considerably on top of the standard RGB baseline, coming close to the benefits from using ground truth depth.
Last, whereas DORN\cite{fu2018dorn} has a lower global RMS error, it leads to notably lower semantic segmentation accuracies.
This confirm our hypothesis that for semantic segmentation it is the local depth structures that are important.
We conclude that estimated depth with D-DFN leads to improved semantic segmentation accuracy.


\subsection{State-of-the-art comparison}

Last, we compare our final model with the state-of-the-art.
For the state-of-the-art comparisons we rely on DFN \cite{yu2018dfn} semantic segmentation networks instead of Deeplab V3~\cite{chen2017deeplabv3}, as DFN yields empirically the best results.
The backbone network is ResNet-101 pre-trained by ImageNet.
We report results in Table~\ref{tab:sota}.

We make the following observations.
For one, the baseline DFN model is close to the top performing RefineNet~\cite{lin2017refinenet}, when considering only RGB channels as the input. 
What is more, when relying on estimated depth for helping semantic segmentation, the proposed 3D Neighborhood convolution manages to get very close to models that must rely on ground truth depth, like \cite{wang2018depthaware} and ours.
This is quite substantial, as the estimated depth is obtained for free with no extra sensory equipment at test time.
Note that \cite{wang2018depthaware} rely on a RefineNet \cite{lin2017refinenet} with an additional pretraining on ADE20K dataset \cite{zhou2016ade} that contains similar indoor scenes, whereas our models do not require extra pretraining.

We also include qualitative results in Figure \ref{fig:qualitative} to show the effectiveness of our 3DN-Conv. We show that a model with depth incorporated by 3DN-Convs outperforms its baseline model in terms of local edge quality, intra-object consistency and high-level semantic classification.

We conclude that 3D Neighborhood convolution can successfully leverage the estimated depth by the proposed D-DFN depth estimation network to improve RGB semantic segmentation, without sensory depth at inference time.

\section{Conclusion}

In this work we introduce depth-aware convolutions around 3D neighborhoods, which adapt the receptive field of convolutions according to the local depth.
Further, we introduce the D-DFN model for estimating depth maps that are locally accurate around semantic boundaries.
As a result, we can now use estimated depth to improve RGB-D based semantic segmentation.
Results on the two datasets validate that indeed using estimated depth we improve semantic segmentation considerably.
We conclude that convolutions that are aware of depth locality and scale improve RGB-D semantic segmentation, even when estimated depth.

\vspace{-5pt}
{\small
\paragraph{Acknowledgment}
This research was supported in part by the SAVI/MediFor and the NWO VENI What \& Where projects.
}


{\small
\renewcommand\refname{References}
\bibliographystyle{ieee}
\bibliography{main.bbl}
\
}

\end{document}